\documentclass[letterpaper]{article} 
\usepackage{aaai2026}  
\usepackage{times}  
\usepackage{helvet}  
\usepackage{courier}  
\usepackage[hyphens]{url}  
\usepackage{graphicx} 
\usepackage{natbib}  
\usepackage{caption} 
\usepackage{algorithm}
\usepackage{algorithmic}

\usepackage{amsthm}
\usepackage{amsfonts}
\usepackage{amsmath}
\usepackage{amssymb}
\usepackage{booktabs}
\newtheorem{definition}{Definition}
\newtheorem{theorem}{Theorem}
\newtheorem{corollary}{Corollary}
\usepackage{newfloat}
\usepackage{listings}
\DeclareCaptionStyle{ruled}{labelfont=normalfont,labelsep=colon,strut=off} 
\floatstyle{ruled}
\newfloat{listing}{tb}{lst}{}
\floatname{listing}{Listing}
\title{Differentially Private Subspace Fine-Tuning for Large Language Models}

\author{
    Lele Zheng\textsuperscript{\rm 1},
    Xiang Wang\textsuperscript{\rm 1},
    Tao Zhang\textsuperscript{\rm 1}\thanks{Corresponding author.},
    Yang Cao\textsuperscript{\rm 2},
    Ke Cheng\textsuperscript{\rm 1},
    Yulong Shen\textsuperscript{\rm 1}
}
\affiliations{
    \textsuperscript{\rm 1}School of Computer Science and Technology, Xidian University, Xi’an, China\\
    \textsuperscript{\rm 2}Department of Computer Science, Institute of Science Tokyo, Tokyo, Japan\\
    \{zhenglele, taozhang, chengke\}@xidian.edu.cn, wxhygge@stu.xidian.edu.cn, \\cao@c.titech.ac.jp, ylshen@mail.xidian.edu.cn
}

\usepackage{bibentry}

\begin{document}

\maketitle

\begin{abstract}

Fine-tuning large language models on downstream tasks is crucial for realizing their cross-domain potential but often relies on sensitive data, raising privacy concerns. 
Differential privacy (DP) offers rigorous privacy guarantees and has been widely adopted in fine-tuning; however, naively injecting noise across the high-dimensional parameter space creates perturbations with large norms, degrading performance and destabilizing training. 
To address this issue, we propose DP-SFT, a two-stage subspace fine-tuning method that substantially reduces noise magnitude while preserving formal DP guarantees. Our intuition is that, during fine-tuning, significant parameter updates lie within a low-dimensional, task-specific subspace, while other directions change minimally. 
Hence, we only inject DP noise into this subspace to protect privacy without perturbing irrelevant parameters. 
In phase one, we identify the subspace by analyzing principal gradient directions to capture task-specific update signals. 
In phase two, we project full gradients onto this subspace, add DP noise, and map the perturbed gradients back to the original parameter space for model updates, markedly lowering noise impact. 
Experiments on multiple datasets demonstrate that DP-SFT enhances accuracy and stability under rigorous DP constraints, accelerates convergence, and achieves substantial gains over DP fine-tuning baselines. 
\end{abstract}

\begin{links}
    \link{Code}{https://github.com/XidianNss/DP-SFT}
\end{links}

\section{Introduction}
\begin{figure}[t]
  \centering
  \includegraphics[width=0.45\textwidth]{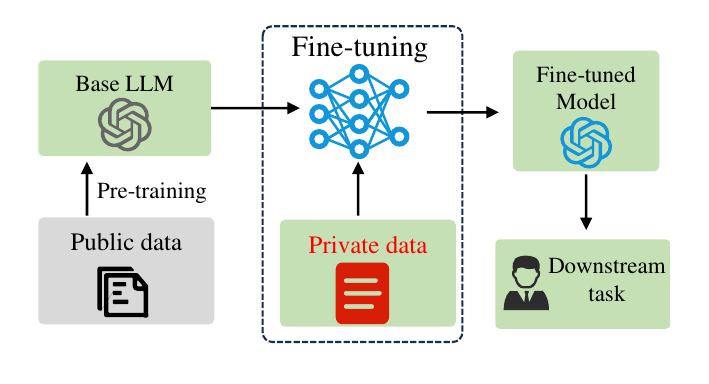}
  \caption{Fine-tuning Process of Large Language Models}
  \label{fig:finetune}
\end{figure}

Large language models (LLMs), such as BERT and GPT, have achieved remarkable success in natural language processing and generation tasks~\cite{devlin2019bert,zhao2023survey,alipour2024chatgpt}. To optimize their performance on specific downstream tasks, fine-tuning is essential~\cite{wang2025parameter,lora,ding2023parameter}. As shown in Fig.~\ref{fig:finetune}, this process allows pre-trained models to be adapted to task-specific data, enhancing accuracy and effectiveness. However, fine-tuning often relies on datasets that contain sensitive information, such as personal identities, financial data, or private conversations. The potential leakage of such information poses serious threats to user privacy and security~\cite{das2025security}. Therefore, balancing privacy protection with the performance of fine-tuning has become a critical challenge.

Differential privacy (DP)~\cite{DP} provides strict privacy guarantees for sensitive data. The core principle of DP is to ensure that an algorithm's output remains consistent despite small changes in a single data sample. In other words, minor adjustments to an individual’s data should not significantly alter the algorithm's results, thus ensuring the protection of user privacy. DP achieves this by introducing noise into the model, making it impossible for external observers to infer any individual data from the model's output. Recently, DP has become widely applied to large model training and fine-tuning, particularly in applications involving sensitive data.~\cite{DP-SGD} introduced DPSGD, which incorporates Gaussian noise into gradients during deep learning, laying the theoretical foundation for privacy protection in the fine-tuning of large language models. Building on this,~\cite{li2024fine} proposed AnaDP, a method that dynamically allocates noise and privacy budgets based on the importance of model parameters during specific training steps. Their research demonstrated that AnaDP performs exceptionally well across multiple datasets, narrowing the performance gap between DP fine-tuning and standard fine-tuning while maintaining privacy. 

Despite significant progress in the application of differential privacy to fine-tuning, existing methods still face the challenge of the curse of dimensionality. To ensure privacy protection, noise must typically be injected into the entire high-dimensional parameter space. This approach increases the noise magnitude substantially, leading to unnecessary performance degradation and training instability, which ultimately affects the model's effectiveness. For instance, given a fixed privacy parameter, the variance of the Gaussian noise added to the gradients increases linearly with the parameter dimensionality. The introduction of high-dimensional noise presents two major challenges. First, the randomness of the noise significantly disturbs each gradient update direction, making the training process more unpredictable and reducing the model's final performance. Second, the added noise negatively impacts the stability of training results. In the absence of noise, the model typically converges gradually over several iterations. However, with the introduction of high-dimensional noise, the training results often exhibit large fluctuations, making it difficult for the model to converge quickly. Therefore, achieving differential privacy protection in large language model fine-tuning without sacrificing model performance remains a significant challenge.

To address the challenges mentioned above, we draw inspiration from subspace differential privacy and propose a DP-SFT mechanism based on gradient subspace optimization. Our approach is founded on the observation that, despite the vast number of parameters in large language models, effective gradient updates during fine-tuning rely heavily on a few dominant gradient directions. These directions form a compact low-dimensional subspace that is significantly smaller than the original parameter space. Consequently, our method first constructs a gradient subspace by extracting the trajectory of full parameter fine-tuning. It then projects the training gradients onto this subspace, adds noise to meet the differential privacy requirements, and finally projects the noisy subspace back into the high-dimensional space to complete the gradient update. Intuitively, DP-SFT reduces the required noise by significantly lowering the dimensionality of the subspace (by several orders of magnitude). Specifically, our contributions are as follows:
\begin{enumerate}
\item We propose Differentially Private Subspace Fine-Tuning (DP-SFT), a novel two-stage framework that injects privacy noise only into a task-specific low-dimensional subspace of gradients. This approach significantly reduces noise magnitude while preserving formal DP guarantees, overcoming the dimensionality challenge in DP optimization for LLMs.

\item We demonstrate subspace transferability across tasks, enabling the subspace to be constructed on public or related datasets without consuming the privacy budget. Our analysis and experiments show that the subspace captures general task geometry rather than private sample-specific information, making DP-SFT practical and privacy-efficient.

\item Extensive experiments demonstrate that DP-SFT achieves near-non-private performance and consistently outperforms strong baselines under different privacy budgets. It sets a new state-of-the-art in privacy-utility trade-offs for differentially private fine-tuning of LLMs.
\end{enumerate}

\section{Related Works}
\label{sec:related}
A straightforward approach to ensuring differential privacy during fine-tuning is to inject noise into gradients, as demonstrated in DP-SGD~\cite{DP-SGD} and DP-Adam~\cite{Adam}. However, due to the curse of dimensionality, directly perturbing gradients in large language models (LLMs) often leads to significant accuracy degradation. To alleviate this issue,~\cite{lowrankDP} proposes a reparameterized gradient perturbation, which reparameterizes each high-rank weight matrix into two low-rank matrices along with a residual matrix. Noise is then added only to the gradients of the low-rank components, and the perturbed gradients are projected back to update the original high-rank weights. However, applying such reparameterization at every update step can introduce training instability.

More recent work~\cite{DPFTLLM} leverages parameter-efficient fine-tuning techniques, such as LoRA~\cite{lora}, Adapter~\cite{Adapter}, and Compacter~\cite{Compacter}, to reduce the number of perturbed parameters. These methods inject noise only into the gradients of additional lightweight plug-in modules, rather than the full model. While this improves efficiency,~\cite{ghostclipping} argues that full-parameter fine-tuning remains essential in many domains. They introduce ghost clipping, a reparameterization-based technique that avoids explicitly instantiating per-example gradients. Nevertheless, existing low-rank reparameterization methods~\cite{repara1, repara2} still require a large number of retained dimensions to maintain accuracy. For example, even after reparameterization, fine-tuning LLaMA-7B still involves updating hundreds of millions of parameters~\cite{repara1}.

In parallel, several approaches in machine learning have been proposed to address the high-dimensional noise problem caused by differential privacy. Some methods~\cite{publicdata1, publicdata2} leverage prior knowledge extracted from public datasets to improve training on private datasets, but they rely on a high degree of similarity between the public and private domains. Other works~\cite{privatelysubspace, subspacelearning} suggest using principal component analysis (PCA) to construct gradient subspaces, thereby reducing noise dimensionality. However, the high computational cost of PCA makes these methods difficult to scale. In contrast, our approach directly extracts the gradient subspace from training dynamics in a computationally efficient manner, enabling accuracy improvements with manageable resource consumption.

\section{Preliminaries}
\label{sec:preliminaries}

\subsection{Differential Privacy}

Differential Privacy (DP)~\cite{DP} enables the analysis of population-level characteristics in a dataset without disclosing information about any individual. This is achieved by adding carefully calibrated noise to statistical queries or to the dataset itself, making it infeasible for an adversary to determine whether a particular individual's data is present. Formally, we present the relevant concepts of differential privacy and its mathematical definition.

\begin{definition}[$(\varepsilon, \delta)$-Differential Privacy]
\label{def:DP}
For a given $\varepsilon \in \mathbf{R}_{\geq 0}$, an obfuscation mechanism $\mathcal{M}$ satisfies $(\varepsilon, \delta)$-DP if and only if for any pair of neighboring datasets $D$, $D^{\prime}$, and any output set $S \in \mathcal{S}$ ($\mathcal{S}$ is the set of all possible outputs), the probability that outputs belong to the same set should be bounded by
\begin{equation}
\Pr(\mathcal{M}(D) \subseteq S) \leq e^\varepsilon \Pr(\mathcal{M}(D^\prime) \subseteq S) + \delta,
\end{equation}
where $\varepsilon$ represents the privacy budget, which is used to measure the degree of privacy protection. A smaller $\varepsilon$ indicates stronger privacy protection but also requires adding more noise. $\delta$ represents the relaxation term, which measures the degree of non-satisfaction of differential privacy to some extent. 
\end{definition}

In practical applications, $\delta$ is usually set to $o\left( \frac{1}{n} \right)$ to ensure the overall protection effect of differential privacy, where $n$ denotes the number of individuals.

Differential privacy exhibits two fundamental and advantageous properties: composition and post-processing invariance. The composition property quantifies the cumulative privacy loss incurred when multiple analyses are performed on the same dataset, enabling rigorous tracking of privacy budgets over time. In contrast, post-processing invariance guarantees that any data-independent computation applied to the output of a differentially private mechanism cannot weaken its privacy guarantees. 

\begin{theorem}[Composition]
\label{def:composition}
Let mechanisms $M_i$ $( i \in \{1, 2, \ldots, T\} )$ satisfy $(\varepsilon_i, \delta_i)$-DP, then their combination $(M_1, M_2, \ldots, M_T)$ satisfies $(\sum_{i=1}^T \varepsilon_i, \sum_{i=1}^T \delta_i)$-DP.
\end{theorem}

\begin{theorem}[Post-processing invariance]
\label{def:post-processing}
Given a deterministic function $f$ and a $(\varepsilon, \delta)$-differentially private mechanism $\mathcal{M}$, then the composite mechanism $f \circ \mathcal{M}$ satisfies $(\varepsilon, \delta)$-DP.
\end{theorem}

\subsection{Gaussian Mechanism in Deep Learning}

In deep learning, the Gaussian mechanism~\cite{GaussianDP1} is widely used to implement differential privacy by adding calibrated noise to gradients during training. Specifically, each sample's gradient \( g \) is first clipped to a predefined norm bound \( C \) to control sensitivity. Gaussian noise with variance determined by \( C \), \( \varepsilon \), and \( \delta \) is then added to the clipped gradients. The noisy gradients are used to update model parameters, enabling privacy-preserving optimization with a balanced trade-off between privacy and utility.

Gopi et al.'s research~\cite{numerical} utilizes Gaussian differential privacy~\cite{GaussianDP2} to provide an effective method for calculating the Gaussian noise that should be added in deep learning. We abbreviate this method as $\mathrm{gdp}(\cdot)$, with the formal usage as follows:

\begin{corollary}[Property of Gaussian noise]
\label{cor:Gaussian}
Given privacy budget $\varepsilon$, relaxation term $\delta$, sampling probability $q$ and training steps $T$, we obtain
\begin{equation}
\sigma = \mathrm{gdp}(\varepsilon, \delta, q, T).
\end{equation}
For clipped gradients of each training data sample, \( \|g\| \leq C \), Gaussian noise with distribution \( \mathcal{N}(0, \sigma^2 C^2 I_d) \) is added to each gradient, where \( I_d \) denotes the \( d \)-dimensional identity matrix and \( d \) is the dimension of the gradient. Under these conditions, each sample is safeguarded by $(\varepsilon, \delta)$-DP.
\end{corollary}

\section{Methods}
\label{sec:method}
\begin{figure*}[t]
    \centering
    \includegraphics[width=1.0\linewidth]{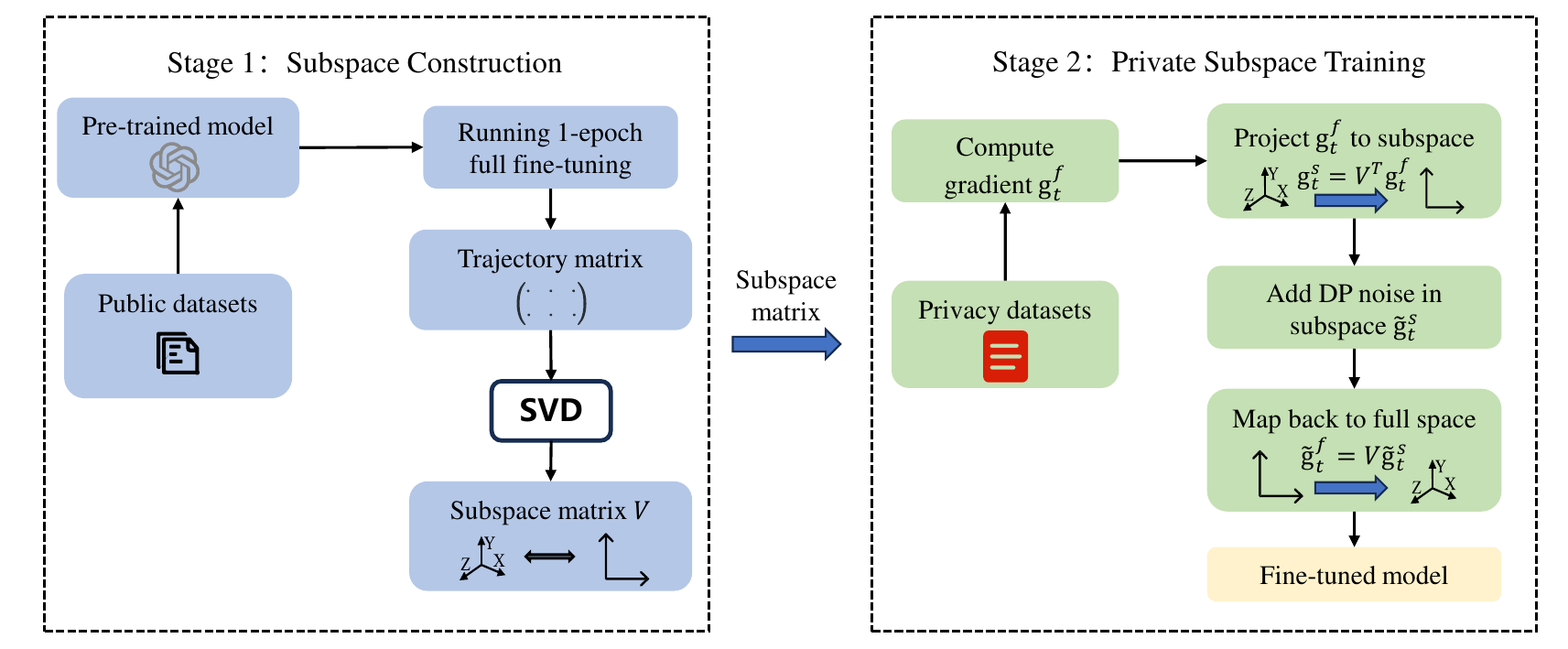}
    \caption{ An overview of the proposed DP-SFT. DP-SFT consists of two stages: (1) Subspace Construction: a set of model snapshots during private training is collected and stacked into a trajectory matrix, from which a task-specific subspace is extracted via SVD. (2) Private Subspace Training: gradients are computed and projected into the subspace, where calibrated DP noise is added before mapping them back to the full parameter space for model updates.
    }
    \label{fig:flowchart}
\end{figure*}

In this section, we present the proposed DP-SFT framework in detail, as illustrated in Fig.~\ref{fig:flowchart}. The goal of DP-SFT is to protect sensitive data used during the fine-tuning of large language models under differential privacy constraints. The core idea of DP-SFT is to significantly reduce the dimensionality of noise injection by restricting it to a task-specific low-dimensional subspace of the gradient space. This strategy improves model utility and training stability while providing formal differential privacy guarantees.

DP-SFT consists of two main stages: Subspace Construction and Private Subspace Training. In the first stage, we identify a low-dimensional subspace that captures the principal directions of task-specific gradient variation. In the second stage, during each training step, incoming gradients are projected onto the identified subspace, Gaussian noise calibrated to satisfy the DP constraint is added, and the perturbed gradients are then mapped back to the original parameter space for model updates.

\subsection{Stage 1: Subspace Construction}
\label{subsec:short-trajectory}

The first stage of DP-SFT aims to construct a task-specific low-dimensional subspace that captures the principal directions of parameter updates during fine-tuning, as shown in Algorithm~\ref{alg:sub-extraction}. Under differential privacy (DP) constraints, directly injecting noise into the full high-dimensional parameter space can severely degrade model performance. To mitigate this, we identify a compact subspace that retains meaningful update directions, thereby improving training stability and model utility.
Similarly, the recent work Mosformer~\cite{cheng2025mosformer} in secure multi-party computation also employs dimensionality reduction strategies to effectively control communication and computational overhead.

Concretely, we perform one epoch of full-parameter fine-tuning on the private dataset $\mathcal{D}$ under $(\varepsilon, \delta)$-DP constraints, and periodically record model parameters. Based on these snapshots, we construct a trajectory matrix:
\begin{equation}
A = \left[ \Theta^{(1)} - \Theta^{(0)}, \Theta^{(2)} - \Theta^{(0)}, \ldots, \Theta^{(m)} - \Theta^{(0)} \right] \in \mathbb{R}^{m \times d},
\end{equation}
where $\Theta^{(j)}$ denotes model parameters at step $j$ $( j \in \{0, 1, \ldots, m\} )$, $d$ is the parameter dimensionality, and $m$ is the number of recorded steps. The matrix $A$ captures the evolution of model weights during early-stage fine-tuning.

We then apply Singular Value Decomposition (SVD) to obtain:
\begin{equation}
A = U \Sigma W^\top,
\end{equation}
where the columns of $W \in \mathbb{R}^{d \times d}$ represent orthogonal directions in the parameter space. We retain the top-$k$ right singular vectors corresponding to the largest singular values and construct an orthogonal projection matrix $P \in \mathbb{R}^{d \times k}$, with $k \ll d$. This projection matrix defines a task-specific subspace that captures the most significant optimization trajectories and serves as a dimensionality reduction operator between the full parameter space $\mathbb{R}^d$ and the subspace $\mathbb{R}^k$.

Importantly, subspace construction requires only short training trajectories to be effective. In our experiments, we further demonstrate that performing SVD on trajectory matrices collected from a brief fine-tuning process is sufficient to yield high-quality subspace projection matrices. These matrices support strong performance during subsequent private training. As a result, the subspace construction stage incurs negligible computational overhead, improving the overall practicality and scalability of our approach.

\begin{algorithm}[t] 
	\caption{Subspace Construction.} 
	\label{alg:sub-extraction} 
	\begin{algorithmic}[1] 
		\REQUIRE 
    Model parameters $\Theta_0$ with $d$ dimensions, subspace dimension $k$, training set $S$, learning rate $\eta$, loss function $\mathcal{L}$, training steps $T$, batch size $B$, clipping threshold $C$, privacy budget $\varepsilon$ and relaxation term $\delta$.
    \ENSURE Trained model parameters $\Theta^*$, Projection matrix $P$.
    \STATE Calculate standard deviation $\sigma$ = gdp$\left(\varepsilon, \delta, \frac{B}{\left\lvert S \right\rvert}, T\right)$; 
    \STATE Calculate trajectory recording period $y = \lfloor \frac{T}{k}\rfloor$;
    \STATE Create an empty matrix $A$;
    \FOR{$t = 1$ to $T$}
        \STATE Sample a batch $\mathcal{B}_t$ of size $B$ from training set $S$;  
        \FOR{$x_i$ in $\mathcal{B}_t$}
            \STATE Calculate  gradients $g_t(x_i)=\nabla_{\Theta_t}\mathcal{L}(\Theta_t, x_i)$;
            \STATE Clip $\overline{g_t}(x_i) = g_t(x_i)/\max\left(1, \frac{\Vert g_t(x_i)\Vert_2}{C}\right)$;
        \ENDFOR
        \STATE Aggregate gradient: $\bar{g}_t \gets \frac{1}{B}\sum_{x_i\in \mathcal{B}_t}\overline{g_t}(x_i)$;
        \STATE Add noise: $\widetilde{g_t} \gets \bar{g}_t + \mathcal{N}(0, \sigma^2 C^2 I_d)$;
        \IF{$t\mid y$}
            \STATE Flatten trajectory $s$ = flat($\Theta_t - \Theta_{0}$);
            \STATE Append $s$ as a new row to matrix $A$;
        \ENDIF
        \STATE $\Theta_{t+1} \gets {Adam}(\Theta_{t}, \widetilde{g_t}, \eta)$;
    \ENDFOR
    \STATE $P \gets \textsc{SVD}(A, k)$
    \RETURN $\Theta^*$, $P$.
	\end{algorithmic}
\end{algorithm}

\subsection{Stage 2: Private Subspace Training}
\label{subsec:stage2}

After constructing the task-specific subspace, DP-SFT performs differentially private fine-tuning by restricting gradient updates to this subspace. The detailed procedure is presented in Algorithm~\ref{alg:sub-opt}. Unlike traditional DP-SGD, which injects noise into the entire high-dimensional parameter space \(g_h \in \mathbb{R}^d\), our method operates solely in the low-dimensional subspace with projected gradients \(g_z \in \mathbb{R}^k\), where \(k \ll d\).

At each training step \(t\), we compute the full per-sample gradient \(g_t(x_i) = \nabla_{\Theta_t} \mathcal{L}(\Theta_t, x_i)\), flatten it, and project it to the subspace using an orthogonal matrix \(P \in \mathbb{R}^{d \times k}\):
\begin{equation}
g_t^s(x_i) = P^\top g_t^f(x_i).
\end{equation}

This projection compresses the gradient into a low-dimensional representation while preserving task-relevant directions.

We then perform \(\ell_2\)-norm clipping with threshold \(C\), followed by Gaussian noise injection:
\begin{equation}
\widetilde{g}_t^s = \text{Clip}(g_t^s, C) + \mathcal{N}(0, \sigma^2 C^2 I_k),
\end{equation}
where \(\sigma\) is calibrated to the target privacy parameters \((\varepsilon, \delta)\). Since noise is injected in \(\mathbb{R}^k\) rather than \(\mathbb{R}^d\) where \(k \ll d\)  (with $k$ being orders of magnitude smaller than $d$), the noise dimension is significantly reduced from $d$ to $k$ in comparison with standard DP-SGD.

The noisy gradient is then mapped back to the original parameter space:
\begin{equation}
\widetilde{g}_t^f = P \cdot \widetilde{g}_t^s \in \mathbb{R}^d,
\end{equation}
reshaped into the original model parameter format, and used to update the model with a standard optimizer such as SGD or Adam.

From the perspective of differential privacy, subspace projection serves as a geometric safeguard. The projection matrix \(P\) effectively filters out gradient components that are orthogonal to the subspace, ensuring that only the most relevant directions are exposed to noise. For any gradient vector $x$, the transformation $P^\top x$ extracts its coordinates in the subspace basis, while $Px$ reconstructs its subspace-aligned component in the original space.

Because the Gaussian mechanism is applied to the clipped gradient in the subspace (with bounded sensitivity), and all subsequent steps such as projection back and parameter reshaping are post-processing operations, the entire training procedure remains \((\varepsilon, \delta)\)-differentially private according to the composition and post-processing invariance properties of differential privacy. Compared to standard DP-SGD, DP-SFT achieves improved training stability and model utility by narrowing the noise injection domain to a low-dimensional, task-relevant subspace.

\begin{algorithm}[t] 
	\caption{Private Subspace Training.} 
	\label{alg:sub-opt} 
	\begin{algorithmic}[1] 
		\REQUIRE 
		Model parameters $\Theta^*$ with $d$ dimensions, matrix $P_{d\times k}$  with column vectors as standard orthogonal basis, training set $S$, learning rate $\eta$, loss function $\mathcal{L}$, training steps $T$, batch size $B$, clipping threshold $C$, privacy budget $\varepsilon$ and relaxation term $\delta$.
		\ENSURE Trained model parameters $\Theta$.
            \STATE Calculate standard deviation $\sigma$ = gdp$\left(\varepsilon, \delta, \frac{B}{\left\lvert S \right\rvert}, T\right)$; 
		\FOR{$t = 1$ to $T$}  
			\STATE Sample a batch $\mathcal{B}_t$ of size $B$ from training set $S$;  
            \FOR{$x_i$ in $\mathcal{B}_t$}
                \STATE Compute gradients $g_t(x_i)=\nabla_{\Theta_t}\mathcal{L}(\Theta_t, x_i)$;
                \STATE Flatten $g_t^f(x_i) = $ flat$\left(g_t(x_i)\right)$;
                \STATE Project $g_t^s(x_i) = P^\top\cdot g_t^f(x_i)$;
                \STATE Clip $\overline{g_t^s}(x_i) = g_t^s(x_i)/\max\left(1, \frac{\Vert g_t^s(x_i)\Vert_2}{C}\right)$;
            \ENDFOR
            \STATE Aggregate gradient: $\bar{g_t^s} \gets \frac{1}{B}\sum_{x_i\in \mathcal{B}_t}\overline{g_t^s}(x_i)$;
            \STATE Add noise: $\widetilde{g_t^s} \gets \bar{g_t^s} + \mathcal{N}(0, \sigma^2 C^2 I_k)$;
    		\STATE Finish projection $\widetilde{g_t^f} = P\cdot\widetilde{g_t^s}$;
            \STATE Recover shape $\widetilde{g_t} = \text{flat}^{-1}\left(\widetilde{g_t^f}\right)$;
            \STATE $\Theta_{t+1} \gets {Adam}(\Theta_{t}, \widetilde{g_t}, \eta)$;
		\ENDFOR   
		\RETURN $\Theta$.
	\end{algorithmic}
\end{algorithm}

\subsection{Privacy Guarantee}
We formally prove that DP-SFT provides $(\varepsilon, \delta)$-differential privacy for each training sample, based on the properties of the Gaussian mechanism and the composition and post-processing invariance theorems of differential privacy.

\begin{theorem}[Differential Privacy of DP-SFT]
\label{thm:dp-sft}
For any $p_1, p_2 \in (0, 1)$ such that $p_1 + p_2 = 1$,   DP-SFT satisfies $(\varepsilon, \delta)$-differential privacy for each data sample $x_i \in \mathcal{D}$.
\end{theorem}

We decompose the training process into two stages and analyze the privacy guarantee of each:

\textbf{(1) Subspace Construction.}  
This phase involves performing a brief phase of full-parameter fine-tuning to extract optimization trajectories. Gaussian noise is added to gradient updates, which satisfies $(p_1 \varepsilon, p_1 \delta)$-DP according to the standard privacy analysis of DP-SGD~\cite{DP-SGD}. As the singular value decomposition (SVD) and subspace construction are deterministic post-processing steps, the post-processing invariance property (Theorem~\ref{def:post-processing}) ensures that this phase remains $(p_1 \varepsilon, p_1 \delta)$-DP.

\textbf{(2) Private Subspace Training.}  
During training, each per-sample gradient $\nabla_{\Theta_t}\mathcal{L}(\Theta_t, x_i)$ is projected into the subspace using $P^\top$, clipped to norm $C$, and perturbed with Gaussian noise drawn from $\mathcal{N}(0, \sigma^2 C^2 I_k)$, where $k \ll d$. According to the Gaussian mechanism~\cite{DP}, this step satisfies $(p_2 \varepsilon, p_2 \delta)$-DP. The reconstruction step using $P$ is again post-processing and does not affect the privacy bound.

By the basic composition theorem~\cite{DP}, the total privacy loss is at most $((p_1 + p_2)\varepsilon, (p_1 + p_2)\delta) = (\varepsilon, \delta)$, completing the proof.

\subsection{Subspace Transferability}
In practice, constructing the task-specific subspace does not require direct access to the private dataset used for downstream fine-tuning. Instead, the subspace can be estimated from a semantically related public dataset by collecting its gradient trajectories. This is feasible because tasks within the same domain often share dominant optimization directions in parameter space.

To verify the aforementioned property, we conduct cross-task transfer experiments in which subspaces are extracted from non-sensitive public datasets (e.g., MNLI or QNLI) without applying differential privacy, and subsequently reused for private fine-tuning on sensitive target tasks. Empirical results indicate that transferring a subspace derived from a different yet related dataset incurs only negligible performance degradation. This behavior suggests that the learned subspace primarily reflects general structural and geometric characteristics of the task family rather than memorizing dataset-specific or sensitive information.

Following established practice in the differential privacy literature, we therefore regard subspace extraction as a public preprocessing step that incurs no privacy cost. The demonstrated transferability further highlights the subspace’s role as a reusable structural prior—both facilitating efficient downstream optimization and ensuring compatibility with privacy guarantees.

\section{Experiments}
In this section, we evaluate the effectiveness of our proposed method, DP-SFT, under different privacy budgets and benchmark tasks. We compare it with strong baselines covering both full-parameter and parameter-efficient differentially private fine-tuning methods. 
\subsection{Experimental Setup}
\label{sec:experiment_settings}
\paragraph{Datasets} To evaluate the effectiveness of DP-SFT, we conduct experiments on four widely used natural language understanding benchmarks: SST-2~\cite{socher2013recursive}, IMDB~\cite{maas2011learning}, QNLI~\cite{wang2018glue}, and MNLI~\cite{williams2017broad}. SST-2 is a binary sentiment classification task consisting of short movie reviews, commonly used to assess fine-tuning stability and sample efficiency. IMDB is another sentiment classification dataset, featuring longer, noisier, and more user-generated reviews, making it more privacy-sensitive. QNLI is a question–answer entailment task derived from the SQuAD dataset, where the model determines whether a candidate sentence correctly answers a given question. And MNLI is a large natural language inference benchmark involving sentence pairs from diverse genres labeled as entailment, contradiction, or neutral. Together, these datasets cover a range of task types and input lengths, enabling a comprehensive evaluation of model utility, robustness, and generalization under strict privacy constraints.

\paragraph{Base Model} All experiments are conducted using the RoBERTa-base model~\cite{liu2019roberta}, a widely adopted and robust pre-trained language model built on the Transformer architecture. RoBERTa-base consists of 12 encoder layers, each with 768 hidden dimensions and 12 self-attention heads, resulting in approximately 125 million parameters. Its strong performance and stability make it a suitable backbone for evaluating differentially private fine-tuning methods across diverse NLP tasks.

\paragraph{Baselines}
To validate the effectiveness of the proposed method, we compare DP-SFT with several representative baselines: 
(1) \textbf{Full-tuning}~\cite{liu2019roberta}, which updates all model parameters without privacy protection; 
(2) \textbf{Full-DP}~\cite{DP-SGD}~, a fully private variant using DP-SGD to ensure \((\varepsilon, \delta)\)-DP; 
(3) \textbf{LoRA-DP}~\cite{yu2021differentially}, a parameter-efficient method applying DP-SGD to low-rank adaptation modules; 
(4) \textbf{Adapter-DP}~\cite{yu2021differentially}, which inserts and trains adapter modules under DP constraints.

\paragraph{Metrics}
To evaluate the effectiveness of differentially private fine-tuning methods, we report accuracy on the validation sets of each dataset. 

\paragraph{Implementation Details} Our implementation is based on HuggingFace’s Transformers library~\cite{wolf2020transformers}. To ensure reproducibility and fair comparison, we use consistent settings across all methods: the batch size is fixed at 32; SST-2 and QNLI have a maximum input sequence length of 128, whereas IMDB and MNLI have a length of 256. For all DP-enabled methods (Full-DP, LoRA-DP, and Adapter-DP), we adopt a learning rate of $5 \times 10^{-4}$ and a clipping threshold of 10. For the proposed DP-SFT, the subspace dimension is set to 32 for SST-2, QNLI, and IMDB, and 64 for MNLI. Privacy parameters are configured as $\delta = 10^{-5}$ for SST-2, QNLI, and IMDB, and $\delta = 10^{-6}$ for MNLI due to their dataset sizes, which is the same as \cite{yu2021large}. The subspace construction stage runs for 1 epoch with a learning rate of $5 \times 10^{-4}$. Privacy parameters are configured as $\delta = 10^{-5}$. We evaluate the methods under two privacy budgets: $\varepsilon \in \{1, 4\}$. More implementation details can be found in the Supp. 1.

\subsection{Experiment Results}
\label{sec:experiment_results}

\paragraph{Performance of DP-SFT}
\label{sec:performance}
To comprehensively evaluate the effectiveness of our proposed framework, we report DP-SFT’s performance under different privacy budgets across four benchmark datasets: SST-2, IMDB, QNLI, and MNLI. As shown in Tables~\ref{tab:final-eps4} and~\ref{tab:final-eps1}, DP-SFT consistently achieves the best performance among all DP methods under both settings. Under $\varepsilon = 4$, DP-SFT outperforms the strongest baseline (Adapter-DP) by up to 11.51\% on QNLI and 11.12\% on MNLI, while maintaining a small gap of only 1.84\% to the non-private full-tuning upper bound. This demonstrates that DP-SFT delivers near-lossless utility even with privacy constraints. Under the more challenging $\varepsilon = 1$ setting, performance degradation becomes more pronounced for all methods, but DP-SFT remains robust. It outperforms Adapter-DP by up to 13.89\% on QNLI and 23.38\% on IMDB, showing strong resistance to utility loss. Remarkably, DP-SFT retains over 86.5\% accuracy on MNLI, while other methods drop below 73\%. These results validate our hypothesis that confining gradient perturbation to a task-specific low-dimensional subspace not only reduces noise impact but also leads to significantly improved model utility and training stability.

\begin{table}[t]
    \centering
    \small
    \begin{tabular}{lcccc}
        \toprule
        \textbf{Method} & \textbf{SST-2} & \textbf{IMDB} & \textbf{QNLI} & \textbf{MNLI} \\
        \midrule
        Full-Tuning (Non-DP) & 0.9507 & 0.9424 & 0.9249 & 0.8777 \\
        Full-DP & 0.7592 & 0.7891 & 0.5995 & 0.3348 \\
        LoRA-DP ($r$=16) & 0.8876 & 0.8886 & 0.8054 & 0.4327 \\
        Adapter-DP ($r$=48) & 0.8909 & 0.8752 & 0.8056 & 0.7566 \\
        \textbf{DP-SFT (Ours)} & \textbf{0.9323} & \textbf{0.9352} & \textbf{0.9207} & \textbf{0.8678} \\
        \bottomrule
    \end{tabular}
    \caption{Accuracy comparison of all methods under standard privacy constraint ($\varepsilon=4$). \textit{Note}: Bold values indicate the best performance among differentially private methods.}
    \label{tab:final-eps4}
\end{table}

\begin{table}[t]
    \centering
    \small
    \begin{tabular}{lcccc}
        \toprule
        \textbf{Method} & \textbf{SST-2} & \textbf{IMDB} & \textbf{QNLI} & \textbf{MNLI} \\
        \midrule
        Full-Tuning (Non-DP) & 0.9507 & 0.9424 & 0.9249 & 0.8777 \\
        Full-DP & 0.7500 & 0.7752 & 0.5854 & 0.3220 \\
        LoRA-DP ($r$=16) & 0.8693 & 0.8394 & 0.7807 & 0.3424 \\
        Adapter-DP ($r$=48) & 0.8624 & 0.6952 & 0.7820 & 0.7242 \\
        \textbf{DP-SFT (Ours)} & \textbf{0.9327} & \textbf{0.9290} & \textbf{0.9209} & \textbf{0.8652} \\
        \bottomrule
    \end{tabular}
    \caption{Accuracy comparison of all methods under extreme privacy constraint ($\varepsilon=1$). \textit{Note}: Bold values indicate the best performance among differentially private methods.}
    \label{tab:final-eps1}
\end{table}

\begin{table}[t]
    \centering
    \small
    \begin{tabular}{lcccc}
        \toprule
        \textbf{Training Scheme} & \textbf{SST-2} & \textbf{IMDB} & \textbf{QNLI} & \textbf{MNLI} \\
        \midrule
        Full-tuning & 0.9507 & 0.9424 & 0.9249 & 0.8777 \\
        4-epoch subspace & 0.9492 & 0.9406 & 0.9232 & 0.8717 \\
        1-epoch subspace & \textbf{0.9438} & \textbf{0.9388} & \textbf{0.9191} & \textbf{0.8665} \\
        \midrule
        \textbf{1-epoch vs Full} & \textbf{-0.69\%} & \textbf{-0.36\%} & \textbf{-0.58\%} & \textbf{-1.12\%} \\
        \bottomrule
    \end{tabular}
    \caption{Accuracy comparison between the few-epoch subspace in stage 1 and full fine-tuning}
    \label{tab:short-trajectory}
\end{table}

\paragraph{Effectiveness of Short Trajectories for Subspace Construction}
To investigate whether short training trajectories can effectively capture task-specific optimization patterns for subspace construction, we perform an ablation study comparing subspace extraction using 1 epoch and 4 epochs of full-parameter fine-tuning, as well as standard 32-epoch full-tuning.
As shown in Table~\ref{tab:short-trajectory}, subspace models trained using only 1 epoch of the trajectory still achieve strong performance. Specifically, the 1-epoch subspace model achieves 94.38\% on SST-2, 93.88\% on IMDB, 91.91\% on QNLI, and 86.65\% on MNLI.
These results demonstrate that short training trajectories (e.g., 1 epoch) suffice to extract meaningful subspaces for effective optimization under privacy constraints. Importantly, this approach reduces subspace construction time by 96.9\%, thereby markedly enhancing the efficiency and applicability of the proposed method while maintaining stability.

\begin{table}[t]
    \centering
    \small
    \begin{tabular}{lcc}
        \toprule
        \textbf{Transfer Direction} & \textbf{Accuracy} & \textbf{$\Delta$ vs Ideal} \\
        \midrule
        IMDB $\rightarrow$ SST-2 & 0.9071 & -3.67\% \\
        SST-2 $\rightarrow$ IMDB & 0.9100 & -2.88\% \\
        \bottomrule
    \end{tabular}
    \caption{Performance of subspace transferability in non-DP settings}
    \label{tab:transfer-non-dp}
\end{table}

\begin{table}[t]
    \centering
    \small
    \begin{tabular}{lccc}
        \toprule
        \textbf{Transfer Direction} & \textbf{Accuracy} & \textbf{$\Delta$ vs Ideal} & \textbf{$\Delta$ vs Full-DP} \\
        \midrule
        IMDB $\rightarrow$ SST-2 & 0.9014 & -4.24\% & +14.22\% \\
        SST-2 $\rightarrow$ IMDB & 0.9090 & -2.98\% & +11.99\% \\
        \bottomrule
    \end{tabular}
    \caption{Performance of subspace transferability under DP ($\varepsilon=4$)}
    \label{tab:transfer-dp}
\end{table}

\paragraph{Subspace Transferability}
In Tables~\ref{tab:transfer-non-dp} and \ref{tab:transfer-dp}, arrows such as ``IMDB $\rightarrow$ SST-2'' denote that the subspace is constructed using the source task (IMDB) and then transferred to the target task (SST-2) for differentially private training. The reported accuracy corresponds to model performance on the target task using the transferred subspace. To evaluate whether subspaces can leverage shared structural patterns for cross-task transfer, we design experiments under both non-private (Non-DP) and differentially private (DP, $\varepsilon = 4$) settings. Our hypothesis is that similar datasets share latent feature distributions, allowing effective subspace reuse without leaking private data.
The results in Table~\ref{tab:transfer-non-dp} confirm that the subspace can encode transferable structure. In Table~\ref{tab:transfer-dp}, DP training with transferred subspaces still maintains high performance, outperforming the Full-DP baseline by 14.22\% on SST-2 and 11.99\% on IMDB, despite small drops compared to ideal training.
These findings verify two key properties: (1) subspaces can generalize across related tasks; (2) transferred subspaces retain model stability and utility under DP noise.
Overall, subspace transfer presents an efficient and privacy-preserving approach to reusing task representations in large language model fine-tuning.

\begin{table}[t]
    \centering
    \small
    \begin{tabular}{lccc}
        \toprule
        \textbf{Dataset} & \textbf{DP-SFT (Noisy)} & \textbf{Full-DP} & \textbf{$\Delta$ vs Full-DP} \\
        \midrule
        SST-2 & 0.7890 & 0.7592 & \textbf{+2.98\%} \\
        IMDB & 0.8596 & 0.7891 & \textbf{+7.05\%} \\
        QNLI & 0.6571 & 0.5995 & \textbf{+5.76\%} \\
        MNLI & 0.3575 & 0.3348 & \textbf{+2.27\%} \\
        \bottomrule
    \end{tabular}
    \caption{Performance degradation from noisy trajectory construction ($\varepsilon_{\text{total}}=4$)}
    \label{tab:noisy-trajectory}
\end{table}

\paragraph{Impact of Noisy Trajectories}
To assess the effect of noise on subspace quality, we design an experiment where training trajectories are generated using DP-SGD with a privacy budget of \(\varepsilon = 3\) in stage 1, followed by low-dimensional DP training with \(\varepsilon = 1\) in stage 2, yielding a total budget of \(\varepsilon_{\text{total}} = 4\).
As shown in Table~\ref{tab:noisy-trajectory}, although DP-SFT (Noisy) consistently outperforms the Full-DP baseline across all datasets, achieving up to +7.05\% improvement on IMDB, its accuracy remains substantially lower than non-private baselines. For example, performance on QNLI (65.71\%) and MNLI (35.75\%) exhibits a sharp decline compared to non-DP references (92.49\% and 87.77\%).
These findings highlight a fundamental limitation: injecting noise during subspace construction significantly reduces the fidelity of the learned subspace in representing the model’s dominant update directions. This motivates our core design principle, which is to construct transferable subspaces using public data. As a result, we avoid early-stage noise contamination while preserving privacy guarantees and improving model utility during subsequent private fine-tuning. We leave the integration with secure inference systems like L-SecNet~\cite{song2024secnet} for future work.

\section{Conclusion}
\label{sec:conclusion}
We proposed DP-SFT, a differentially private subspace fine-tuning framework that mitigates the curse of dimensionality in LLM optimization under privacy constraints. By restricting noise injection to a task-specific low-dimensional subspace extracted via SVD on short training trajectories, DP-SFT substantially reduces the amount of noise required for privacy while preserving formal $(\varepsilon, \delta)$-DP guarantees. Extensive experiments across multiple NLP benchmarks demonstrate that DP-SFT consistently outperforms existing full and parameter-efficient DP fine-tuning methods in both accuracy and training stability. Furthermore, we show that the extracted subspaces are transferable across tasks and can be constructed with negligible computational overhead, offering a scalable and practical solution for privacy-preserving adaptation of large models.

\section*{Acknowledgments}
This paper is supported by the National Natural Science Foundation of China (62572373, 62220106004, 62402358, 92467201, 62502363, U24A20238), the National Key R\&D Program of China (2023YFB3107500), the Key R\&D Program of Shandong Province of China (2023CXPT056, 2025CXPT089), the Young Talent Fund of Association for Science and Technology in Shaanxi, China (20240138), the Natural Science Basic Research Program of Shaanxi Province (2025JC-YBQN-869), the Aeronautical Science Foundation of China (20181981006), the Open Topics from the Lion Rock Labs of Cyberspace Security (\#LRL24004), the Fundamental Research Funds for the Central Universities (ZDRC2202, ZYTS25081, KYFZ25005), the Xidian University Specially Funded Project for Interdisciplinary Exploration (TZJHF202502), the China Scholarship Council (CSC), the Double First-Class Overseas Research Project of Xidian University, and JSPS KAKENHI JP23K24851, JST PRESTO JPMJPR23P5, JST CREST JPMJCR21M2, JST NEXUS JPMJNX25C4.

\bibliography{aaai2026}

\end{document}